\documentclass{article}

\usepackage[margin=1in]{geometry}

\usepackage{microtype}
\usepackage{graphicx}
\usepackage{subfigure}
\usepackage{booktabs} 
\usepackage{amsmath}
\usepackage{amsfonts}
\usepackage{enumitem}
\usepackage{amsthm}
\usepackage{algorithm2e}
\usepackage{algorithmic}

\usepackage{todonotes}
\usepackage{comment}

\usepackage{tikz}

\theoremstyle{definition}

\usepackage{hyperref}

\begin{document}

\title{Learning Large DAGs by Combining Continuous Optimization and Feedback Arc Set Heuristics}

\author{Pierre Gillot and Pekka Parviainen\\University of Bergen, Norway}
\date{}
\maketitle

\begin{abstract}
Bayesian networks represent relations between variables using a directed acyclic graph (DAG). Learning the DAG is an NP-hard problem and exact learning algorithms are feasible only for small sets of variables. We propose two scalable heuristics for learning DAGs in the linear structural equation case. Our methods learn the DAG by alternating between unconstrained gradient descent-based step to optimize an objective function and solving a maximum acyclic subgraph problem to enforce acyclicity. Thanks to this decoupling, our methods scale up beyond thousands of variables.
\end{abstract}

\section{Introduction}

Bayesian networks are probabilistic graphical models that represent joint distributions among random variables. They consist of a structure which is a directed acyclic graph (DAG) representing conditional independencies and parameters that specify local conditional distributions.

Bayesian networks can handle both discrete and continuous variables. In this work, we concentrate on continuous variables. Specifically, we study linear structural equation models (SEMs) where the local conditional distribution in a node is a Gaussian whose mean is linear combination of the values of its parents. 

Traditionally, there are two main approaches for learning DAGs. In {\em constraint-based} approach (see, e.g., \cite{pearl00, spirtes00}), one performs conditional independence tests and tries to construct a DAG that expresses the same conditional independencies as the test results. We take the {\em score-based} approach (see, e.g., \cite{Cooper1992, Heckerman1995}) where one tries to find a DAG that maximizes a score. Typically, one uses decomposable scores, that is, the score of a DAG is a sum of local scores for each node-parent set pair. This leads to a combinatorial optimization problem where one picks a parent set for each node while satisfying the constraint that the resulting graph is acyclic. 

The combinatorial learning problem is NP-hard \cite{chickering96} and developing scalable methods is challenging. Indeed, state-of-the-art exact learning methods scale only up to few hundred nodes \cite{cussens11} and scalable algorithms for SEMs rely on approaches such as local modifications \cite{Aragam_2019}. A recent breakthrough, NOTEARS \cite{zheng2018dags} circumvents the combinatorial problem by formulating a continuous acyclicity constraint. This enables usage of gradient-based optimization methods. The bottleneck with respect to scalability lies in the cubic complexity for the calculation of the acyclicity function which involves the computation of a matrix exponential. GOLEM \cite{NEURIPS2020_d04d42cd} is similar to NOTEARS but replaces the generalized LASSO objective found in NOTEARS by a log-likelihood-based fitness function. It shares however the same computational bottleneck as NOTEARS due to the acyclicity constraint. Some methods circumvent this bottleneck by finding a sparse graph without the acyclicity constraint and impose acyclicity afterwards \cite{DBLP:journals/corr/abs-2006-03005, yu20}.

Our goal is to develop a fast heuristic for learning linear DAGs. In other words, we trade accuracy for speed. We speed-up learning by decoupling the optimization of the objective function from the acyclicity constraint in a similar fashion as \cite{DBLP:journals/jmlr/ParkK17}\footnote{Differences are discussed in Section~\ref{sec:method}.}. At a general level, we learn by iteratively repeating the following steps:
\begin{enumerate}
    \item Given an acyclic graph, find a graph (possibly cyclic) which is better in terms of the objective function value.
    \item Given a cyclic graph, find an acyclic graph.
\end{enumerate}

The first step can be solved efficiently using state-of-the-art gradient-based solvers. We present two variants for this step. {\em ProxiMAS} uses proximal gradient descent whereas {\em OptiMAS} uses standard automatic differentiation and gradient-based updates.

In the second step, the cyclic solution from the first step is converted into an acyclic one. The quality 
of the final solution depends crucially on the quality of this conversion. We solve an instance of maximum acyclic subgraph (MAS) problem which has been previously used to learn DAGs \cite{pmlr-v138-gillot20a}. Intuitively, we prefer keeping arcs whose weights are far from zero. Solving the MAS problem exactly is NP-hard but there exists efficient heuristics for solving its complement, the feedback arc set (FAS) problem \cite{10.14778/3021924.3021930}.

Our experiments (Section~\ref{sec:exp}) show that our methods can quickly find reasonable solutions on datasets with thousands variables, even on modest running time. OptiMAS and ProxiMAS perform well compared to GD, NOTEARS and GOLEM in large-scale learning when resources are limited in terms of processors and memory.

\section{Background}

\subsection{Linear structural equation models and Bayesian network structure learning} \label{sec:sem}

A Bayesian network is a representation of a joint probability distribution. It consists of two parts: a structure and parameters. The structure of a Bayesian network is a DAG $G=(N, A)$ where $N$ is the node set and $A$ is the directed adjacency matrix; we denote the parent set of a node $v$ by ${Pa}_v$. Parameters specify local conditional distributions $ P(v\mid {Pa}_v)$ and the joint probability distribution is represented by a factorization
\[P(N) = \prod_{v\in N} P(v\mid {Pa}_v).\]

We consider linear structural equation models (SEMs) where local conditional distributions are Gaussian distributions whose mean is a linear combination of the values of the parents of the variable. 
The structure of a linear SEM is determined by a weight matrix $W\in \mathbb{R}^{d\!\times\!d}$; $W(i, j)$ is non-zero if and only if $A(i,j)=1$, that is, there is an arc from $i$ to $j$. For a $d$-dimensional data vector $x$, we have 
\[
x = xW + e,
\]
where $e$ is a $d$-dimensional noise vector. The elements of $e$ are independent of each other.

The goal in Bayesian network structure learning is to find a DAG $G$ that fits the data. We are given a data matrix $X\in \mathbb{R}^{n\!\times\!d}$ with $n$ samples of $d$-dimensional vectors. Our goal is to find a weight matrix $W\in \mathbb{R}^{d\!\times\!d}$ that represents an acyclic graph. To quantify how well the DAG and the weights fit the data, we can use the least-squares loss $\frac{1}{2n}\|XW-X\|_2^2$. Furthermore, we want to induce sparsity and therefore we add a regularization term $g(W)$. This leads to the following optimization problem.
\begin{equation} \label{eq:sem}
\begin{aligned}
\underset{W}{\text{argmin}} \text{ } & \frac{1}{2n}\|XW-X\|_2^2 + \lambda_1 g (W)\\
\text{s.t.} \quad & W \text{ is acyclic.}
\end{aligned}
\end{equation}
In the above formulation, $\lambda_1$ is a user-defined constant that determines the strength of regularization. To induce sparsity, we regularize with $L1$-norm, that is,  $g(W)= \|W\|_1 = \sum_{i,j} |W(i, j)|$.

\subsection{Maximum acyclic subgraph and feedback arc set} \label{sec:mas}

Formally, the maximum acyclic subgraph (MAS) problem is defined as follows. We are given a directed graph $G = (V, E)$ and a weight function $w(e)$ that assigns a weight for each arc $e\in E$. The goal is to find an acyclic graph $G' = (V, E')$ such that $E'\subseteq E$ and $\sum_{e\in E'} w(e)$ is maximized.

The maximum acyclic subgraph problem has a dual (or complementary) problem:  the feedback arc set (FAS) problem. In FAS, we are given a directed graph $G = (V, E)$ and a weight function $w(e)$ just like in MAS. The goal is to find an arc set $E''$ such that $G'' = (V, E\setminus E'')$ is acyclic and $\sum_{e\in E''} w(e)$ is minimized. It is well-known that $E' = E\setminus E''$. Thus, MAS can be solved by first solving FAS and then performing a simple subtraction of sets to obtain the solution to MAS. 

Both MAS and FAS are NP-hard \cite{Kar72}. Therefore, exact algorithms are intractable on large graphs. Fortunately, there exists fast heuristics for FAS \cite{berger90, EADES1993319, 10.14778/3021924.3021930}.

\section{Proposed method} \label{sec:method}

A critical difficulty in solving Equation \ref{eq:sem} stems from a combination of two problems:
\begin{itemize}
    \item The quadratic objective function for the linear SEM problem has at most $nd^3$ quadratic terms. Indeed, the quadratic expression $\|XW-X\|_2^2$ is a sum of $n\!\times\!d$ squared expressions $\big((XW)_{i,j}-X_{i,j}\big)^2$, where each $(XW)_{i,j}$ is a linear expression consisting of $d$ terms. As $X$ is a continuous data matrix, one can rarely simplify the quadratic objective function significantly.
    \item Enforcing acyclicity. Standard constraints lead to NP-hard combinatorial problems. In the continuous setting, a smooth function exists that encodes acyclicity but with a prohibitive cubic complexity.
\end{itemize}
The main contribution of this work therefore is to address these two concerns. First, we decompose the quadratic optimization problem into a sequence of easier subproblems using iterative optimization techniques. Second, we separate entirely the quadratic optimization from the acyclicity constraints. Acyclicity is enforced by solving a MAS task as a proxy. The outline of the proposed method is shown in Algorithm \ref{alg:proximas} which iteratively does the following steps at each iteration $k$:
\begin{enumerate}
    \item A new objective function is created based on the acyclic solution $W_{k-1}$ obtained at the end of the previous iteration, which penalizes the original linear SEM objective by the least-square term $\frac{\lambda_2}{2} \|W - W_{k-1} \|_2^2$.
    \item An optimization step is performed on the MAS-penalized problem, leading to a new cyclic solution $\widetilde{W}_k$.
    \item An acyclic projection $W_k$ of the previously obtained cyclic solution $\widetilde{W}_k$ is extracted, based on the squared values of $\widetilde{W}_k$. Formally, we attempt to compute $W_k = \widetilde{W}_k \odot A_k$, where $A_k$ is the solution of the following MAS problem:
\begin{equation} \label{eq:MAS}
\begin{aligned}
A_k &= \underset{A}{\text{ argmax }} {\displaystyle\sum_{i,j}}|\widetilde{W}_k(i,j)|^2 A(i,j)\\
&\quad\quad\quad\text{s.t.} \quad A \in \{0,\!1\}^{d\!\times\!d} \text{ is acyclic.}
\end{aligned}
\end{equation}
    As mentioned in Section~\ref{sec:mas}, finding optimal solutions for MAS is usually too time consuming and one has to resort to heuristics. We use a vectorized version of the approximation algorithm by Eades \cite{EADES1993319} to find the acyclic weighted adjacency matrix $W_k$ (Algorithm \ref{alg:greedy}).
\end{enumerate}
Intuitively, steps 1-2 are designed such that the updated weights matrix $\widetilde{W}_k$ will be constrained to remain in the vicinity of the previously found acyclic solution $W_{k-1}$ returned by the MAS heuristic.  In that sense, we approximate the acyclicity function used in NOTEARS and GOLEM by a projection term toward acyclic solutions which is much easier to compute and differentiate. Step 3 aims to preserve edges that represent the most important dependencies. In other words, we want to keep the weights that are far from zero. Solving MAS using weights that are squares of the original weights is equivalent to minimizing $\|W_k - \widetilde{W}_k\|_2$, which corresponds to finding the acyclic solutions that are closest to the cyclic solutions returned by the iterative optimization process. By alternating between optimization steps and MAS extractions via the repetition of steps 1-3, we aim to navigate through the search space of the original linear SEM problem by ``following the trail'' of a sequence of dynamically generated acyclic solutions. 

The GD algorithm introduced in \cite{DBLP:journals/jmlr/ParkK17} follows a similar strategy. It proceeds by repeating the three following steps: 1) make a gradient step for the linear SEM objective, 2) project the current cyclic solution to its MAS solution and 3) fit the linear SEM problem constrained by the newly found acyclic structure; as an optional fourth step, when the progress is too small they resort to a swapping order heuristic. The main difference compared to our method is that we do not perform steps 3 and 4. The GD algorithm is greedier than our method, since we do not attempt to optimize the parameters of every discovered acyclic structure. From a practical standpoint, at each step of the GD algorithm, $d$ LASSO instances have to be solved which becomes intractable for large-scale structure learning. In comparison, by directly plugging in step 1 the MAS projections to the linear SEM objective as dynamically evolving penalization terms, our approach circumvents entirely the need to solve these LASSO instances. 

\subsection{Connection with online convex optimization}

Perhaps surprisingly, the dynamic nature of the proposed optimization procedure is not particularly challenging to work with in practice. Algorithm \ref{alg:proximas} can, indeed, be seen as a special case of an online convex optimization (OCO) problem. In his seminal paper \cite{10.5555/3041838.3041955}, Zinkevich introduces this framework which he defines as such:
\begin{itemize}
    \item $F \subset \mathbb{R}^n$ is a feasible set (assumed bounded, closed and non-empty).
    \item $(c_k)_k$ is an infinite sequence of smooth convex functions from $F$ to $\mathbb{R}$, with bounded gradients.
    \item At each step $k$, an element $x_k \in F$ is selected, then assigned the cost $c_k(x_k)$.
\end{itemize}
In OCO, the standard optimization error becomes ill-defined and one seeks to optimize instead the so-called regret defined as 
\begin{equation*}
    \text{regret } = \text{ } \sum_{k\leq K} c_k(x_k) - \min_{x \in F} \sum_{k \leq K} c_k(x).
\end{equation*}
Zinkevich  was the first to extend the gradient descent algorithm to its online form. It is well known that assuming convexity of the $c_k$ and boundedness of the gradients, online gradient descent achieves $\mathcal{O}(\sqrt{K})$ regret bound and this bound is improved to $\mathcal{O}\big(\log(K)\big)$ assuming strong convexity of the $c_k$ \cite{DBLP:journals/corr/abs-1909-05207}. More general classes of OCO algorithms have been studied \cite{NIPS2009_ec5aa0b7, DBLP:journals/corr/abs-1810-03594}, notably (accelerated) proximal gradient descent algorithms concerned about composite convex functions of the form $\phi_k = f_k + g$ where only the $f_k$ are smooth. Improved regret bounds again hold assuming strong convexity of the $\phi_k$.

The proposed method is therefore theoretically well-behaved: by considering the functions $f_k: W \mapsto \frac{1}{2n}\|XW-X\|_2^2 + \frac{\lambda_2}{2} \|W - W_{k-1} \|_2^2$ and $\phi_k: W \mapsto f_k(W) + \lambda_1 g(W)$ (Algorithm \ref{alg:proximas} line 2), notice that every $\phi_k$ is $\lambda_2$-strongly convex since for each $k$, the function $W \mapsto \phi_k(W) - \tfrac{\lambda_2}{2}\|W - W_{k-1} \|_2^2=\frac{1}{2n}\|XW-X\|_2^2 + \lambda_1 g(W)$ is convex; Algorithm \ref{alg:proximas} therefore inherits aforementioned regret bounds from the OCO setting.

\subsection{Implementation details}

We implemented two variants of the proposed method:
\begin{itemize}
    \item The first implementation, {\bf ProxiMAS}, is designed to take full advantage of the properties of the objective functions $\phi_k$, owing to the decoupling with acyclicity. Recall that we have $f_k: W \mapsto \frac{1}{2n}\|XW-X\|_2^2 + \frac{\lambda_2}{2} \|W - W_{k-1} \|_2^2$ and $\phi_k: W \mapsto f_k(W) + \lambda_1 g(W)$, where $g$ is convex and the $f_k$ are smooth and convex. By smooth, we mean that every $f_k$ is differentiable with its gradient defined as $\nabla f_k(W) = \tfrac{1}{n}X^t X (W - I) + \lambda_2 (W-W_{k-1})$ and using the Cauchy-Schwarz inequality, one can easily show that every $f_k$ has a Lipschitz-continuous gradient with optimal Lipschitz constant $L_k$ upper bounded by $L = \tfrac{1}{n}\|X^tX + n\lambda_2I\|_2$, a value that does not depend on $k$. We can therefore use a proximal gradient descent optimization scheme as a backbone for our implementation, hence the name ProxiMAS. In practice, we use the FISTA algorithm \cite{FISTA}, an accelerated proximal algorithm with $\mathcal{O}(\tfrac{1}{k^2})$ convergence rate (in an offline optimization setting). One should notice that the running time of ProxiMAS does not depend on the number of samples $n$, since the proximal updates depend only on the covariance matrix $X^tX \in \mathbb{R}^{d\!\times\!d}$ which can be pre-computed.
    \item The second implementation, {\bf OptiMAS}, replaces the proximal gradient descent by gradient descent-like steps. The main interest in doing so is that automatic differentiation will handle the optimization using a generic gradient descent-based solver. Despite the linear SEM objective being non-differentiable when the regularization term is the $L1$ norm, automatic differentiation frameworks can in practice optimize such non-smooth objective. Thus, OptiMAS is agnostic to the choice of the optimizer. In principle, one can use any variant of gradient-based optimizers. In our implementation, we have used Adam \cite{DBLP:journals/corr/KingmaB14}. 
\end{itemize}
We stress that both variants are taking full advantage of vectorization and are thus GPU accelerated, first because we lifted the need to solve a sequence of LASSO instances at each step, second because the MAS heuristic (Algorithm \ref{alg:greedy}) is efficiently vectorized and runs quasi-linearly with respect to the number of nodes $d$ when a GPU is available.

\begin{figure}

\begin{minipage}[t]{0.47\textwidth}
\begin{algorithm}[H] 
	\caption{Proposed method}
	\label{alg:proximas}
	\begin{center}
	\begin{algorithmic}[1]
	    \REQUIRE Data $X \!\in\! \mathbb{R}^{n\!\times\!d}$, initialization $W_0 \!\in\! \mathbb{R}^{d\!\times\!d}$, number of iterations $K$, $\lambda_1>0$, $\lambda_2 > 0$, optimizer
	    \ENSURE Approximate solution to Equation~\ref{eq:sem}
		\FOR {$1 \leq k \leq K$} 
		    \STATE Define $f_k: W \mapsto \frac{1}{2n}\|XW-X\|_2^2 + \frac{\lambda_2}{2} \|W - W_{k-1} \|_2^2$ $\phi_k: W \mapsto f_k(W) + \lambda_1 g(W)$
		    \STATE Make an optimization step on $\phi_k$: $\widetilde{W}_k = \text{step}(\phi_k, \text{ optimizer})$
		    \STATE Project updated weights to their MAS approximation: $W_k = \text{greedy\_MAS}\big(\widetilde{W}_k\big)$
	    \ENDFOR
	    \STATE {\bfseries return} $W_{\text{best}} = \underset{k}{\text{ argmin }} \tfrac{1}{2n} \|XW_k-X\|_2^2 + \lambda_1 g(W_k)$
	    \texttt{\vspace{0.3cm}}
	\end{algorithmic} 
	\end{center}
\end{algorithm} 
\end{minipage}
\hfill
\begin{minipage}[t]{0.47\textwidth}
\begin{algorithm}[H] 
	\caption{Vectorized greedy MAS}
	\label{alg:greedy}
	\begin{center}
	\begin{algorithmic}[1]
	    \REQUIRE $\widetilde{W} \!\in\! \mathbb{R}^{d\!\times\!d}$
	    \ENSURE Approximate solution to Equation \ref{eq:MAS}
        \STATE $\hat{W} = \widetilde{W} \odot \widetilde{W}$
        \STATE $\text{scores} = \hat{W}\!.\text{sum}(\text{dim}\!=\!0)$
        \STATE $\text{order} = \text{zeros}(\text{size}\!=\!d)$
        \STATE $\text{ub} = (d+1)\times \max(\text{scores})$
		\FOR {$0 \leq i < d$} 
		    \STATE $\text{node} = \text{ argmin }\text{scores}$
		    \STATE $\text{order}[-(i+1)] = \text{node}$
		    \STATE $\text{scores}[\text{node}] = \text{ub}$
		    \STATE $\text{scores} = \text{scores} - \hat{W}[\text{node},:]$
	    \ENDFOR
	    \STATE $\text{order}^{-1} = \text{ argsort }\text{order}$
	    \STATE $W = \text{lower\_triangular}\big(\widetilde{W}[\text{order},\text{order}]\big)$
	    \STATE $W = W[\text{order}^{-1},\text{order}^{-1}]$
	    \STATE {\bfseries return} $W$
	\end{algorithmic} 
	\end{center}
\end{algorithm} 
\end{minipage}

\end{figure}

\section{Experiments} \label{sec:exp}

We now present our experimental pipeline. We choose to compare the proposed algorithms (ProxiMAS and OptiMAS) against an iterative method (GD \cite{DBLP:journals/jmlr/ParkK17}) and the current state-of-the-art methods for sparse linear SEM structure recovery (NOTEARS \cite{zheng2018dags} and GOLEM 
\cite{NEURIPS2020_d04d42cd}).

\subsection{Data generation}

We adopt a similar setup as in \cite{zheng2018dags, NEURIPS2020_d04d42cd}: we first generate random DAGs based on Erd{\H o}s-R\'enyi ("ER") and scale-free ("SF") models. We consider three sparsity regimes: sampled DAGs have $k\times d$ edges, where $d$ is the number of nodes and $k \in \{1, 2, 4\}$. Graphs are denoted by "ERk" or "SFk" depending on their graph model and sparsity. Then, we generate the weighted adjacency matrices $W$ by assigning random weights uniformly sampled in the range $[-2,-0.5] \cup [0.5, 2]$. Finally, we generate samples $X$ following the linear SEM model: $X=E (I-W)^{-1}$, where $E \in \mathbb{R}^{n\!\times\!d}$ represents $n$ i.i.d. samples from either a Gaussian, exponential or Gumbel distribution in $\mathbb{R}^d$. For all distributions, we investigate both the equal variance ("EV") setting with scale $1.0$ for all variables and the non-equal variance ("NV") setting where every variable has its scale sampled uniformly in the range $[0.5, 1.5]$. Unless stated otherwise, $n$ samples are generated both for the training data and for the validation data, with $n \in \{1000, 10000\}$. 

\subsection{Metrics}

In order to evaluate the performance of the different methods, we compute the false negative rate (FNR), false positive rate (FPR) and the normalized structural Hamming distance (SHD) between predicted and groundtruth adjacency matrices. We proceed similarly with the undirected adjacency matrices. The Gaussian negative log-likelihood is also computed on the validation data (unseen during training). Aforementioned metrics are extracted at different thresholding values of the predicted weights matrices. Different methods behave differently at a fixed thresholding value. For example, we observed in our large-scale tests with limited running time that, for any fixed threshold, OptiMAS tend to produce significantly sparser graphs than NOTEARS and GOLEM. Thus, OptiMAS has lower FPR and higher FNR. In order to get a general performance score independent from the choice of a thresholding value, we additionally consider the average precision score as implemented in the scikit-learn package. This metric is robust against strong class imbalance as it occurs in large-scale sparse structure recovery. For brevity, only a fraction of the figures are shown in this paper. 

\subsection{Implementation} \label{sec:implementation}

The two proposed methods (ProxiMAS and OptiMAS) are implemented in pytorch 1.8. The GOLEM method comes in two variations GOLEM-EV and GOLEM-NV originally implemented in tensorflow. In order to streamline benchmarking these variations were re-implemented in pytorch. The tensorflow and pytorch implementations were compared at fixed seed and produce nearly identical results given the same data; speedwise, we found the difference between the two implementations to be insignificant for large scale graphs with thousands of nodes. The original implementation of NOTEARS relies on an L-BFGS-B solver implemented in scipy and as mentioned in \cite{NEURIPS2020_d04d42cd} it does not scale to large instances with thousands of variables, thus for fairness we re-implemented it in pytorch as well. The existing implementation of the GD algorithm is written in R and uses the highly optimized package glmnet, thus we did not alter the implementation. All methods have full GPU support, with the exception of GD which relies on the LASSO implementation from the glmnet package and is restricted to CPU.  For equal comparison, all methods are tested in a multi-threaded setting, without GPU.

\subsection{Hyperparameters}

All the tested methods have a hyperparameter $\lambda_1$  that regulates sparsity. An additional hyperparameter $\lambda_2$ exists for the GOLEM, NOTEARS ProxiMAS and OptiMAS methods which enforces "dagness". The values of these two hyperparameters yield different behavior depending on the method. The chosen value of $\lambda_1$ for NOTEARS and the chosen values of $\lambda_1$ and $\lambda_2$ for GOLEM are those recommended by their authors. The original NOTEARS implementation is based on an augmented Lagrangian method and does not use the $\lambda_2$ hyperparameter. We added this hyperparameter to our pytorch implementation of NOTEARS the same way as in the GOLEM implementation. We do not claim to have performed any model selection, but chose values that worked well in our tests. The table below gives the values of the hyperparameters depending on the method:
\begin{center}
\begin{table}[!htbp]
\caption{\label{tab:hyperparam}Sparsity and dagness hyperparameters for each method}
\begin{tabular}{ |c|c|c|c|c|c|c| } 
\hline
 & OptiMAS & ProxiMAS & NOTEARS \cite{zheng2018dags} & GOLEM-EV \cite{NEURIPS2020_d04d42cd} & GOLEM-NV \cite{NEURIPS2020_d04d42cd} & GD \cite{DBLP:journals/jmlr/ParkK17} \\
\hline
$\lambda_1$ & $0.1$ & $0.1$ & $0.1$ & $0.02$ & $0.002$ & $0.1$ \\
$\lambda_2$ & $20.0$ & $20.0$ & $5.0$ & $5.0$ & $5.0$ & -\\
\hline
\end{tabular}
\end{table}
\end{center}
Additionally, in all experiments the ProxiMAS and OptiMAS methods are configured to start enforcing acyclicity after $50$ minutes of solving time, whereas NOTEARS and GOLEM enforce acyclicity the entire time as in their original papers. Finally, the methods that rely on automatic differentiation (NOTEARS, GOLEM and OptiMAS) all use the Adam optimizer \cite{DBLP:journals/corr/KingmaB14} as a backbone for gradient descent, with default learning rate $0.001$ as in \cite{NEURIPS2020_d04d42cd}.

\subsection{Benchmarking pipeline}\label{sec:pipeline}

The experiments were run on a cluster with Intel Xeon-Gold 6138 2.0 GHz / 6230 2.1 GHz CPUs. The number of cores and amount of memory used in each experiments are shown in Table~\ref{tab:param}.

We present three different experiments to emphasize the advantageous scaling of the proposed methods comparatively to the state of the art. Table~\ref{tab:param} below shows the different instances we ran for every experiment.

\begin{center}
\begin{table}[!tbph]
\caption{\label{tab:param}Parameters for each experiment}
\begin{tabular}{ |l|l|l|l| } 
\hline
 & Experiment 1 & Experiment 2 & Experiment 3\\
\hline
 $d$ & $1000, 5000$ & $5000$ & $5000, 10000, 15000, 20000$\\
 $k$ & $1, 2, 4$ & $1$ & $1$ \\
 $n$ & $1000, 10000$ & $1000, 10000$ & $1000, 10000$\\
 Graph type & ER, SF & ER & ER\\
 Noise type & Gaussian, exponential, Gumbel & Gaussian & Gaussian\\
 Scale type & EV, NV & EV, NV & EV\\
 Repetitions & $10$ & $1$ & $1$ \\
 Total instances & $1440$ & $4$ & $8$ \\
 CPU cores & $4$ & $4$ & $32$\\
 Memory (GB) & $16$ & $16$ & $128$ \\
 Runtime (h) & $1$ & $24$ & $1$\\
 \hline
\end{tabular}
\end{table}
\end{center}

\subsection{Results}

\begin{center}
\begin{figure}[!htpb]
\begin{tikzpicture}[scale=1]
\node[inner sep=0pt] at (0,0)
    {\includegraphics[scale=0.45]{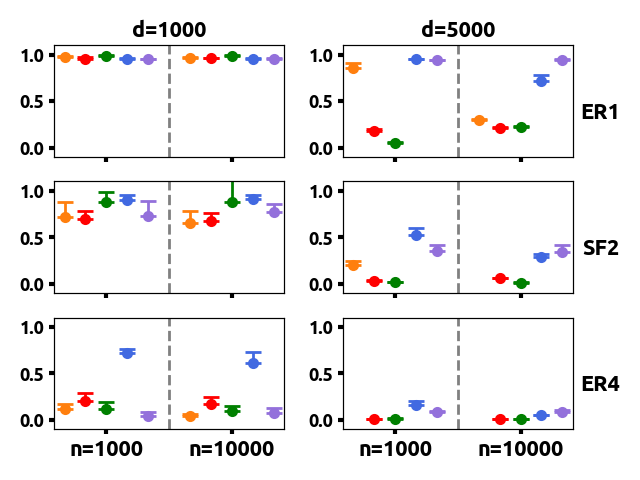}};
\node[inner sep=0pt] at (7.25,0)
    {\includegraphics[scale=0.45]{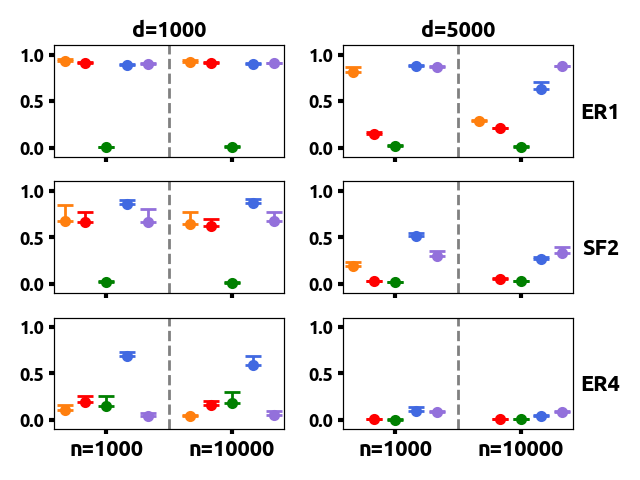}};
\node at (-0.1,3) {Gaussian-EV};
\node at (7.15,3) {Gaussian-NV};
\node[inner sep=0pt] at (0,-6.25)
    {\includegraphics[scale=0.45]{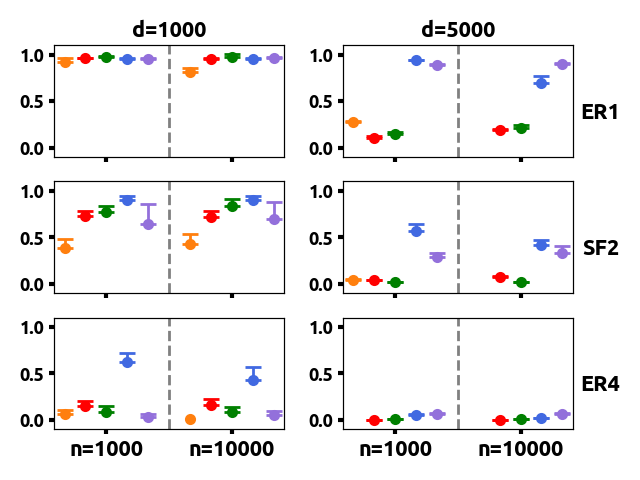}};
 \node[inner sep=0pt] at (7.25,-6.25)
    {\includegraphics[scale=0.45]{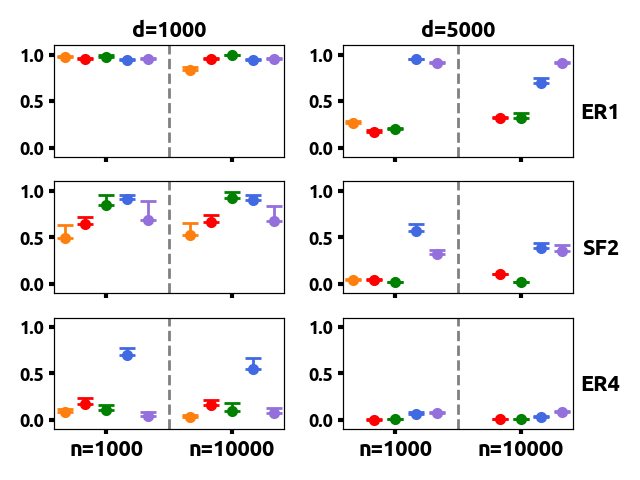}};   
\node[inner sep=0pt] at (3.625,-9.25)
    {\includegraphics[scale=0.6]{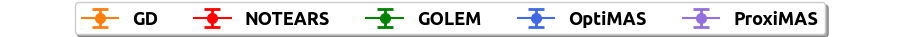}};
 \node at (-0.1,-3.25) {Exponential-EV}; 
 \node at (7.15,-3.25) {Gumbel-EV};  
\end{tikzpicture}
\caption{\label{fig:exp1}Average precisions for different noise distributions and data set sizes. $d$= number of nodes, $n$= number of samples. Confidence intervals show the standard deviation over $10$ data sets.}
\end{figure}
\end{center}

In the first experiment (Experiment~1 in Table~\ref{tab:param}), we generated data with different noise models. We show selected results in Figure~\ref{fig:exp1}.

Overall, OptiMAS and ProxiMAS overperform the benchmark methods. Generally, GD performs equally good as OptiMAS and ProxiMAS when $d=1000$ or $n=1000$. However, it becomes slow when data sets grow. Especially, GD usually fails to even find a solution when there are lots of samples ($n=10000$). Most of the time, NOTEARS and GOLEM are in par with GD or slightly better. Comparing OptiMAS and ProxiMAS, we notice that their performance is usually similar to each other. The main difference is that OptiMAS performs significantly better on more complex graphs (ER4) with $d=1000$; We suspect that this difference is due to numerical instabilities.

\begin{figure}[!htpb]
\begin{center}
\begin{tikzpicture}[scale=1]
\node[inner sep=0pt] at (0,-4.5)
    {\includegraphics[scale=0.6]{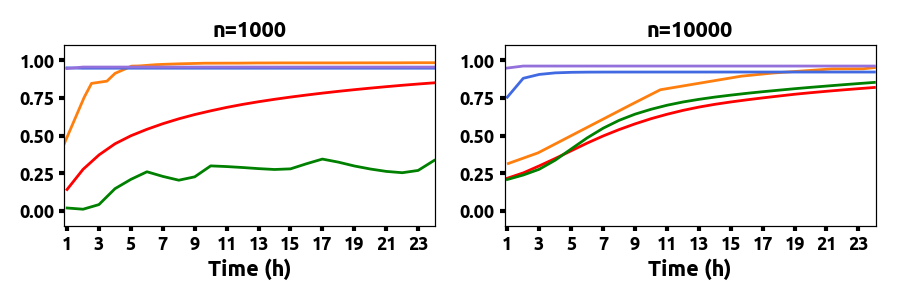}};
\node[inner sep=0pt] at (0,-7)
    {\includegraphics[scale=0.6]{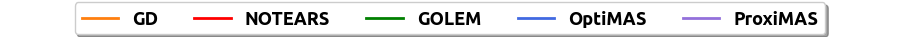}};
\end{tikzpicture}
\caption{\label{fig:exp2}Average precision measured at different time points. Data generated from ER1 with $5000$ nodes and Gaussian-EV noise. Note that on the plot on the left hand side, the curves for OptiMAS and ProxiMAS are overlapping.}
\end{center} 
\end{figure}

We also wanted to analyze how the available running time affected each method. Therefore, we generated two data sets, one with $1000$ samples and the other with $10000$ samples, from an ER1 model with Gaussian-EV noise (Experiment~2 in Table~\ref{tab:param}) and let all methods run for 24 hours. We recorded a snapshot of the weight matrix $W$ every hour. Average precisions are shown in Figure~\ref{fig:exp2}.

We observe that both OptiMAS and ProxiMAS find good solutions quickly. However, improvement after the first hour is negligible. GD starts slowly but eventually catches up with OptiMAS and ProxiMAS and often ends up with a slightly better solution. As in Experiment~1, we notice that the scalability of GD suffers from having lots of observations. Initially, NOTEARS is far behind but keeps improving significantly afterwards and after 24 hours it has found a solution that is almost as good as the ones found by OptiMAS and ProxiMAS. GOLEM performs similarly with NOTEARS when there are $10000$ samples but struggles with $1000$ samples. 

\begin{center}
\begin{figure}[!htpb]
\begin{tikzpicture}[scale=1]
\node[inner sep=0pt] at (0,0)
    {\includegraphics[scale=0.6]{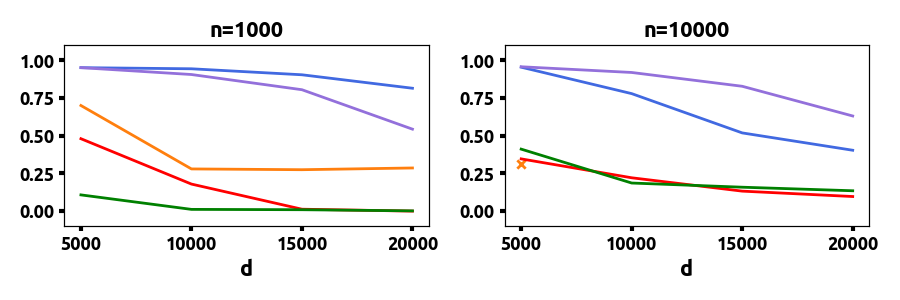}};
\node[inner sep=0pt] at (0,-2.5)
    {\includegraphics[scale=0.6]{fig/legend_lines.png}};
\end{tikzpicture}
\caption{\label{fig:exp3}Scalability of different methods. Average precision is measured with different number of nodes $d$. Data was generated from ER1 with Gaussian-EV noise.}
\end{figure}
\end{center}

Next, we study the scalability of the different methods. To this end, we varied the number of nodes between $5000$ and $20000$ and generated either $1000$ or $10000$ data samples from an ER1 model with Gaussian-EV noise (Experiment~3 in Table~\ref{tab:param}). All methods were given 1 hour running time. Average precisions are shown in Figure~\ref{fig:exp3}.
We notice that with $1000$ samples the average precision for OptiMAS is high for all data set sizes and decreases only little when the number of nodes grows. However, with $10000$ samples average precision drops faster when the number of nodes grows. This may seem counter-intuitive as one would expect that more observations would lead to better performance. The likely explanation for this behavior is that, due to the fixed running time, OptiMAS performed fewer iterations and this countered the effect of increase number of observations at this scale. We can contrast this behavior to ProxiMAS whose running time does not depend on the number of observations. With $n=1000$, ProxiMAS starts with nearly as high average precision as OptiMAS but its performance deteriorates quickly after $10000$ nodes. However, with $n=10000$, the drop is less significant and ProxiMAS clearly outperforms OptiMAS when there are $10000$ or more nodes. We also notice that GOLEM and NOTEARS struggle to learn anything within an hour when there are more than $10000$ nodes. GD performs better than GOLEM and NOTEARS when $n=1000$ but when $n=10000$ it only finds a solution for $d=5000$.

Table~\ref{tab:mem_time} shows memory usage of the different methods and their time per iteration (with acyclicity enforced for NOTEARS, GOLEM, OptiMAS and ProxiMAS). We see that OptiMAS is clearly most memory-efficient. ProxiMAS uses more memory than OptiMAS but much less than NOTEARS and GOLEM. The memory consumption of GD is larger than ProxiMAS but smaller than NOTEARS and GOLEM. Time per iteration for GD is very inconsistent due to the order swapping heuristic it uses at certain iterations, thus we omitted it; as a rule, we observed that GD scales unfavorably with respect to both number of nodes and number of samples.

\begin{center}
\begin{table}[!htpb]
\caption{\label{tab:mem_time}Estimation of the memory usage and time per iteration (32 cores, ER1, Gaussian-EV, n=1000)}
\begin{tabular}{ |l|c|c|c|c||c|c|c|c| } 
\hline
 & \multicolumn{4}{|c||}{Memory (GB)} & \multicolumn{4}{|c|}{Time/iteration (s)} \\
 \cline{1-9}
 $d$ & $5000$ & $10000$ & $15000$ & $20000$ & $5000$ & $10000$ & $15000$ & $20000$ \\
\hline
 OptiMAS & $1$ & $6$ & $13$ & $24$ & $1$ & $2$ & $4$ & $6$ \\
 ProxiMAS & $1$ & $7$ & $14$ & $25$ & $1$ & $3$ & $7$ & $13$ \\
 NOTEARS \cite{zheng2018dags} & $6$ & $23$ & $52$ & $92$ & $6$ & $40$ & $100$ & $250$ \\
 GOLEM \cite{NEURIPS2020_d04d42cd} & $6$ & $23$ & $53$ & $94$ & $6$ & $45$ & $120$ & $280$ \\
 GD \cite{DBLP:journals/jmlr/ParkK17} & $4$ & $12$ & $27$ & $47$ & $-$ & $-$ & $-$ & $-$ \\
 \hline
 \end{tabular}
 \end{table}
 \end{center}

\section{Discussion} \label{sec:discussion}

We presented two different heuristics (ProxiMAS and OptiMAS) for the structure recovery problem in the linear SEM case, revolving around a decoupling of the acyclicity constraints from the continuous optimization itself. We observed that both methods have excellent scaling (both space and time). OptiMAS scales particularly well when the number of samples $n$ is small. On the contrary, ProxiMAS has invariant scaling with respect to $n$ and scales in practice better than OptiMAS when the number of samples is large.

In our observations, both ProxiMAS and OptiMAS tend to get stuck on local extremum: the sequence of acyclic DAGs returned by the two methods is conditioned by the initial cyclic solution provided to them. This drawback can be alleviated by ``warm-starting'': run the algorithm initially without the MAS penalization and extraction steps (Algorithm \ref{alg:proximas} lines 2 and 4), then add these steps at some point during the execution. This strategy is made viable since a single MAS step is enough to guarantee acyclicity. Our experiments show that in practice, very good DAGs can be found even when most of the running time is dedicated to fitting the model without enforcing acyclicity.

Based on our experiments, OptiMAS and ProxiMAS are most competitive in situations where there is a large number of nodes and limited amount of computational resources. If there are a couple of thousands of nodes or less, the current state of the art is preferred. Similarly, if one can afford to run GD, NOTEARS or GOLEM for a long enough period of time, these algorithms will eventually outperform ProxiMAS/OptiMAS. However, in such a situation one could use OptiMAS or ProxiMAS to find an initial solution and use it to ``warm-start'' GD, NOTEARS or GOLEM.

Another limitation of our methods is that it is unclear at the moment how the theoretical results from online convex optimization translate with respect to the original problem. Currently, we are not aware of any necessary condition for the convergence of the proposed methods. This opens up an avenue of future research: Can we prove anything about the quality of the solutions? Can we say something for a specific type of data? Does the fact that we use a heuristic to find a maximum acyclic subgraph has an effect and would improving MAS also translate in better structure learning?

\subsection*{Acknowledgements}

The computations were performed on resources provided by 
UNINETT Sigma2 - the National Infrastructure for High Performance Computing and 
Data Storage in Norway.

The authors thank Young Woong Park for providing the R code for GD.

\bibliographystyle{plain}
\bibliography{bibliography}

\end{document}